\documentclass[11pt,a4paper]{article}
\usepackage{times}
\usepackage{mathptmx}
\usepackage{latexsym}

\usepackage{url}

\usepackage{graphicx}

\title{Modular Mechanistic Networks: On Bridging Mechanistic and Phenomenological Models with Deep Neural Networks in Natural Language Processing}

\author{Simon Dobnik\\
CLASP and FLOV\\
University of Gotenburg, Sweden\\
\texttt{simon.dobnik@gu.se} \\\and
John D. Kelleher\\ 
ADAPT Centre for Digital Content Technology\\
Dublin Institute of Technology, Ireland\\
\texttt{john.d.kelleher@dit.ie}
}

\date{}

\begin{document}
\maketitle

\begin{abstract}
  Natural language processing (NLP) can be done using either top-down (theory driven) and bottom-up (data driven) approaches, which we call \emph{mechanistic} and \emph{phenomenological} respectively. The approaches are frequently considered to stand in opposition to each other. Examining some
  recent approaches in deep learning we argue that deep neural networks
  incorporate both perspectives and, furthermore, that leveraging this aspect of deep learning may help in solving complex problems within language technology, such as modelling language and perception in the domain of spatial cognition.
\end{abstract}

\section{Introduction}
\label{sec:introduction}

There are two distinct methodologies to build computational models of
language or of world in general. The first approach
can be characterised as qualitative, symbolic and driven by domain
theory (we will call this a \emph{top-down} or \emph{mechanistic approach}), 
whereas the second approach may be characterised as quantitative, numeric
and driven by data and computational learning theory (we will call this the \emph{bottom-up} or \emph{phenomenological approach}). In this context we are borrowing the terminology of \emph{phenomenological model} from the literature on the Philosophy of Science where the term \emph{phenomenological model} is sometimes used to describe models that are independent of theory (see for example \cite{mcmullin:1968}), but more generally is used to describe models that focus on the observable properties (phenomena) of a domain (rather than explaining the hidden mechanisms relating these phenomena) \cite{frigghartmann:2017}. For this paper we use the term phenomenological model to characterise models which are primarily driven by fitting to observable relationships between phenomena in a domain, as represented by correlations between features in a dataset sampled from the domain; as opposed to models that are derived from a domain theory of the interactions between domain features.  The focus of this paper
is to examine and frame the potentially synergistic relationship
between these distinct analytic methods for natural language
processing (NLP) in the light of recent advances in deep neural networks (DNNs) and deep learning. 

In historic terms this discussion is recurrent
throughout the history of NLP. For example, early approaches such as \cite{Shieber1986,Alshawi:1992aa} 
are mechanistic in nature as they are based on logic and other formal approaches such as features structures and unification which are tools that allow formalisation of domain theories. With the availability of large corpora in mid-1990s there was a shift to data-driven phenomenological approaches with a focus on statistical machine learning methods \cite{ManningSchutze:1999,Turney:2010aa}. This inspired several discussions on
the relation between the two approaches (e.g., \cite{Gazdar:1996aa,Jones:2000aa}). We share the view of some that both approaches are in fact
in a complimentary distribution with each other as shown in Table~\ref{tab:rules-vs-data} (adapted from a slide by Stephen Pulman). Mechanistic approaches provide deep coverage but of a limited domain;
outside a domain they prove brittle and therefore limited. On the
other hand, phenomenological approaches are wide-coverage and
robust to variation found in data but provide a shallow representation of language.

\begin{table}[htpbp]
\begin{center}
\begin{tabular}{|r|c|c|}
  \hline
  \emph{tech/cov} & wide & narrow \\
  \hline
  deep & \textbf{our goal} & symbolic \\
  \hline
  shallow & data-based & \textbf{useless} \\
  \hline
\end{tabular}
\end{center}
\caption{Properties of mechanistic and phenomenological approaches in NLP}\label{tab:rules-vs-data}
\end{table}

Our desiderata is a wide-coverage system with deep analyses. It was
considered that this could be achieved by a hybrid model but working
out such a model has proven not a trivial task. Systems that
used both approaches treated them normally as independent black-boxes organised
in layers (e.g. \cite{Kruijff:2007}). However, the marked recent
advances in the NLP based on \emph{deep} (!) neural networks
have made the question of how these two methodologies should be used,
related and integrated in NLP research apposite.



The choice of a method depends on the goal of the task for which it is
used. One goal for processing natural language is to develop useful
applications that help humans in their daily life, for example machine
translation and speech recognition.  In application scenarios where a
rough analysis is acceptable (e.g., a translation that provides the
gist of the message) and large annotated and structured corpora are
available, machine learning is the methodology of choice to address
this goal. However, where precise analysis is required or where there
is a scarcity of data, a machine learning approach may not be
suitable. Furthermore, if the goal of processing language is rather
motivated by the desire to better understand its cognitive
foundations, than a machine learning methodology, particularly one
based on an unconstrained, fully connected deep neural network, is
not appropriate. The criticisms of unconstrained neural network based
models (typically characterised by fully-connected feed-forward
multi-layer networks) in cognitive science has a long history (see
\cite{massaro:1988} \emph{inter alia}) and often focuses on (i) the
difficultly in analysing in a domain-theoretic sense how the model
works, and (ii) the, somewhat ironic, scientific short-coming that
neural networks are such powerful and general learning mechanisms that
demonstrating the ability of a network to learn a particular mapping
or a function is scientifically useless from a cognitive science
perspective. In particular, as Massaro \cite{massaro:1988} argues, a neural
network model is so adaptable that given the appropriate dataset and
sufficient time and computing power it is likely to be able to learn
mappings that not only support a cognitive theory but also ones that
contradict that theory. One approach to address this problem is to
introduce domain relevant structural constraints into the model via
the network architecture, early approaches include
\cite{feldman:1988,Feldman1989,regier:1996}. Indeed, we argue in this
paper that one of the important and somewhat overlooked factors
driving the success of research in deep learning is the specificity
and modularity of deep learning architectures to the tasks they are
applied too.

\paragraph{Contribution:} In this paper we evaluate the relation between mechanistic and phenomenological models and
argue that although it appears that the former have lost their
significance in computational linguistics and its applications they are still very
much present in the form of formal language modelling that underlines
most of the current work with machine learning. Moreover, we highlight that many of the recent advances in deep learning for NLP are not based on unconstrained neural networks but rather that these networks have task specific architectures that encode domain-theoretic considerations. In this light, the relationship between mechanistic and phenomenological models can be viewed as potentially more synergistic. Given that many logical theories are defined in terms of \emph{functions} and \emph{compositional} operations and neural networks learn and compose functions, a logic-based domain theory of linguistic performance can naturally inform the structural design of deep learning architectures and thereby merge the benefits of both in terms of model interpretability and performance.

\paragraph{Overview:} 
In Section \ref{sec:deeplearning}, we discuss recent developments in
deep learning approaches in NLP and situate them within
the current debate; then, in Section \ref{sec:spatlang}, we use the
computational modelling of spatial language as an NLP case study to
frame the possible synergies between formal models and machine
learning and set out our thoughts for potential approaches to
developing a more synergistic understanding of the formal models 
and machine
learning for NLP research. In Section \ref{sec:future} we give our
concluding thoughts.

\section{Deep Learning: A New Synthesis?}
\label{sec:deeplearning}



In recent years deep learning (DL) models have improved or in some cases markedly improved the state of the art across a range of NLP tasks. Some of the drivers of DL success include: (i) the availability of large datasets, (ii) more powerful computers, and (iii) the power of learning and adaptability of connectionist neural networks. However, another and less obvious driver of DL is the fact that (iv) DL network models often have architectures that are specifically tailored or structured to the needs of a specific domain or task. This fact becomes obvious when one considers the variety of DL architectures that have been proposed in the literature. For example, a schematic overview of neural network architectures can be found at at: \url{http://www.asimovinstitute.org/neural-network-zoo/} \cite{van-Veen:2016aa}.  


\subsection{Modularity in Deep Learning Architectures}

There are a large-number of network design parameters that may be driven by experimental results rather than domain theory. For example, (i) the size of the network, (ii) the depth of the layers, (iii) the size of the matrices passed between the layers, (iv) activation functions and (v) optimiser are all network parameters that are often determined through an empirical trial-and-error process that is informed by designer intuition \cite{Jozefowicz:2016aa}. However, the diversity of current network architectures extends beyond differences in these parameters and this diversity of network architecture is not a given. For example, given the flexibility of neural networks, one approach to accommodating structure into the processing of a network is to apply minimal constraints on the architecture and to rely on the ability of the learning algorithm to induce the relevant structure constraints by adjusting the network's weights. 

On the other hand, it has, however, long been known that pre-structuring a neural network by the careful design of its architecture to fit the requirements of the task results in better generalisation of the model beyond the training dataset \cite{LeCun:1989aa}. Understood in this context, DL is
assisted (or supervised!) by the task designer in terms of a priori background knowledge who decides
what kind of networks they are going to build, the number of layers, what kind of layers, the connectivity between the layers
and other parameters. 
DL is most frequently not using fully connected layers, instead
several kinds of layered networks have been developed tailored to the
task. In this respect DL models capture top-down domain informed specification that we have
seen with the rule-based NLP systems. This flexibility of neural networks is ensured by their modular design which takes as a basis a single perceptron unit which can be thought of encoding a simple concept. When several units are organised and connected into larger collections of units, these may be given interpretations that we give to symbolic representations in rule-based systems. The level of conceptual supervision may thus vary from no-supervision when fully connected layers are used, to weak supervision that primes the networks to learn particular structures, to strong supervision where the structure is given and only parameters of this structure are trained.

An example of weak supervision are Recurrent Neural Networks (RNNs)
that capture sequence learning required for language models. The design of current state-of-the-art RNN language models is informed by linguistic phenomena such as short- and long-distance dependencies between linguistic units. In order to improve the ability of RNNs to model long-distance dependencies, contemporary RNN language models use Long-Short Memory Units (LSTM) or Gated Recurrent Units (GRUs) which may be further augmented with attention mechanisms \cite{salton:2017}. The inputs and outputs of such networks can be either characters or words, the latter represented as
word embeddings in vector spaces. 

Another example of weakly supervised neural networks, in the sense that their design is informed by a domain, are Convolutional Neural Networks (CNNs) which have their origin in image processing \cite{LeCun:1989aa}. In CNNs the convolutions are meant as filters that encode a
region of pixels into a single neural unit which learns to respond to the occurrence of a pixel pattern in the region specific visual feature. Importantly, the weights associated with a specific convolution are shared across a group of neurons such that together the group of neurons check for the occurrence of the visual features across the full surface of the image.  Additionally, as objects or entities may occur in different parts of image, to decrease
the effects of spatial continuum, operations such as pooling are used
that encode convolved representations from various parts of the
image. In analogy to learning visual features, CNNs have also been used for language modelling to capture different patterns of characters in strings \cite{kim:2016}. 

\begin{figure*}
\includegraphics[width=\textwidth]{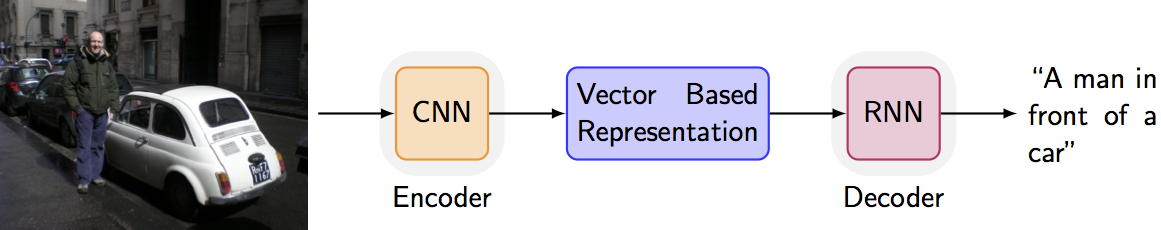}
\caption{A schematic representation of DL image captioning architectures}
\label{fig:image-captioning-schematic}
\end{figure*}

Specialised networks may be treated as modules which are sequenced after each other. For example, the current Neural Machine Translation (NMT) architecture is the encoder-decoder \cite{NIPS2014_5346,bahdanau:2015,luong:2015,kelleher:2016}. This architecture uses one RNN, known as the encoder, to fully process the input sentence and generate its vector based representation. This is passed to a second RNN, the decoder, which implements a language model of the target language which generates the translation word by word. Domain theoretic considerations have affected the design how the two language modelling networks are connected in a number of ways. For example, an understanding that different languages have different word orders lead to enabling the decoder to look both back and forward along the input sentence during translation. This is implemented by fully processing the input sequence with the first RNN before translation is generated by the second RNN. However, the understanding of the need for local dependencies between different sections of the translation and somewhat a contrary requirement to the need for a potentially global perspective on the input has resulted in the development of attention mechanisms within the NMT framework. This means that DL network architectures modules are not only sequenced but they are also stacked. A variant of the NMT encoder-decoder architecture that replaces the encoder RNN with a CNN has revolutionised the field of image captioning \cite{Xu:2015aa}. Figure~\ref{fig:image-captioning-schematic} gives a schematic representation of such image captioning systems. The CNN module learns to represent images as vector representations of visual features and the RNN module is a language model whose output is conditioned on the visual representations. We have already mentioned that CNNs are also used to generate word representations. These representations are then passed to an RNN model to predict the next word in the context of preceding words in the sequence (see \cite{kim:2016}). The advantage of using a CNN module to learn word representation is that it enables the system to capture spelling variation of morphologically-rich languages or texts from social media that does not use standard spelling of words. This and also the preceding examples therefore illustrate how different levels of linguistic representations are modelled in modular DL architectures. 

In summary, the design of a DL architectures, where DL networks are
treated as composable modules, can constrain and guide a number of
factors that are important in representing language and other
modalities, in particular the hierarchical composition of features and
the sequencing of the representations. Importantly, the neural
representations that are used in these cases are inspired by rich work
on top-down rule-based mechanistic natural language processing.

\subsection{Phenomenological versus Mechanistic Models}

The ability to treat neural networks as composable modules within an overall system architecture is a powerful one. This is because during training it is possible to back-propagate the error through each of the system's modules (networks) and train them in consort while permitting each module to learn its distinctive task in parallel with the other modules in the network. However, the power of this approach has led to some research being based on a relatively shallow understanding of domain theory and most of the work being spent on fitting the hyper-parameters of the training algorithm through a grid-search driven by experimental performance on gold-standard datasets. The domain theory is only used to inform the broad outlines of the system architecture. Using image-captioning as an example, and at the risk of presenting a caricature, this approach may be described as: ``we are doing image-captioning so we need a CNN to encode the image and an RNN to generate the language and we will let the learning algorithm sort out the rest of the details''.

This theory free, or at least, theory light approach to NLP research is primarily driven by performance on gold-standard datasets and lamentably frequently the analysis of the systems is limited to the presentation of system results relative to a state-of-the-art leader-board with relatively little reflection on the how the structure of the model reflects theoretic considerations. This focus on performance in terms of accurately modelling the empirical relationship between inputs and outputs and where the trained model is treated as a black box aligns with what we describe as the \emph{phenomenological tradition} in machine learning. This can be contrasted with an alternative tradition within machine learning which is sometimes described as being based on \emph{mechanistic models}. Mechanistic models presuppose a domain theory and the model is essentially a computational implementation of this domain theory. To illustrate this difference, contrast for example the approach to training a support vector machine classifier where multiple kernels are tested until one with high performance on a dataset is found versus the approach to defining the topology of a Bayesian network in such a way that it mirrors a theory informed model of the causal relationships between relevant variables in the domain \cite{kelleher2015fundamentals}. Once the theoretical model has been implemented, the free parameters of the model can then be empirically fit to the data. 


Consequently, mechanistic models are informed by both top-down theoretical considerations of a task designer but they are also sensitive to bottom-up empirical considerations, the training data. Mechanistic models have several advantages, for example: they can be used to test a domain theory. If the model is accurate, this provides evidence that the theory is correct. Assuming the theory is correct, they are likely to outperform phenomenological models in contexts where data is limited.\footnote{See discussion on generative versus discriminative models in \cite{kelleher2015fundamentals}.} 
 The top top-down approach provides background knowledge that restricts the size of the training search space. 

Traditionally, neural networks have been considered the paradigmatic example of a phenomenological model. However, viewing neural networks as component modules within a larger deep-learning systems opens the door to sophisticated mechanistic deep-learning models. Such an approach to network design is, however, dependent on the system designer being informed by domain theory and is therefore strongly supervised in terms of background knowledge. 
An example of modular networks where each module is some configuration of neural units that are tailored to optimise parameters of a particular task is described in \cite{Andreas:2016aa} who work in the domain of question
answering. The architecture learns how to map questions and visual or
database representations to textual answers. In order to answer a question, the
network learns a network layout of modules that are responsible for
the individual steps required to answer the question. For example, to
answer ``What colour is the bird'' the network applies the attention
module to find the object from the question, followed by a module that
identifies the colour of the attended region in the image. The
possible sequences of modules are constrained by being represented as
typed functions: in fact the modules translate to typed functional
applications through which compositionality of linguistic meaning is
ensured as in formal semantics \cite{BlackburnBos:2005}. The system
learns (using reinforcement learning) a layout model which predicts
the sequence of modules to produce an answer for a question sentence
and an execution module which learns how to ground a network layout in the
image or database representation. An extension of this work is
described in \cite{Johnson:2017aa} where both procedures rely on less background knowledge. For example, the system does not use a dependency
parser to parse the input sentence but an LSTM language module and
the modules use a more generic architecture.

The modular networks are in line with the  \emph{structured connectionism} of \cite{feldman:1988} and \emph{constrained connectionism} of  Regier \cite{regier:1996} ``in which complex domain-specific structures are built into the network, constraining its operation in clearly understandable and analysable ways'' \cite[p.~2]{regier:1996}. Regiers's presentation of constrained connectionism is based on a case study on learning spatial relations and events. The case study describes the design and training of a neural network that receives short movies of 2 two-dimensional objects, a static rectangle and a circle which is either static or moving, as input and the model learns to predict the correct spatial term to describe the position 
and movement of the circle relative to the rectangle. For example, a static circle might be described as \emph{above} the rectangle, whereas a moving circle might move \emph{out from under} 
the rectangle. A crucial aspect of this case study for Regier's argument is that the neural network's architecture is constrained in so far as it incorporates a number of structural devices that are motivated by neurological and psychological evidence concerning the human visual system, including motion buffers, angle and orientation computations components, and boundary and feature maps for objects in the input. Following \cite{regier:1996}, in the next section we will take spatial language as an NLP case-study and discuss how domain theory can be used to extend current deep-learning systems so as to move them further towards the mechanistic pole within the phenomenological versus mechanistic spectrum.




\section{Spatial Language}
\label{sec:spatlang}

Our focus is computational modelling of spatial language, such as
\emph{the chair is to the left and close to the table} or \emph{go
  down the corridor until the large painting on your right, then turn
  left}, which requires integration of different sources of knowledge
that affect its semantics, including: (i) scene geometry, (ii)
perspective and perceptual context, (iii) world knowledge about
dynamic kinematic routines of objects, and (iv) interaction
between agents through language and dialogue and with the environment through perception. Below we describe these properties in
more detail:


\emph{Scene geometry} is described within a two-dimensional or
three-dimen\-sional coordinate frame in which we can represent
locations of objects as geometric shapes as well as angles and
distances between them. Over a given area we can identify different
degrees of applicability of a spatial description, for example with
spatial templates \cite{LoganSadler:1996,Dobnik:2017ac}.  A spatial
template may be influenced by perceptual context through the presence
of other objects in the scene known as distractors
\cite{kelleher2005context,costellokelleher:06}, occlusion
\cite{kelleher/vanGenabith:2006,kelleher:2011}, and attention \cite{RegierCarlson:2001}.

Directionals such as \emph{to the left of} require a model of
\emph{perspective} or \emph{assignment of a frame of reference}
\cite{Maillat:2003} which includes a viewpoint parameter. The
viewpoint may be defined linguistically \emph{from your view} or
\emph{from there} but it is frequently left out. Ambiguity with
respect to the intended perspective of a reference can affect the
grounding of spatial terms in surprising ways
\cite{radvanskyLogan:1997,kelleher2005cognitive}. However, frequently
the intended perspective can be either inferred from the perceptual
context (if only one interpretation is possible, see for example the
discussion on contrastive versus relative meanings in
\cite{kelleher2005contextNLG}) or it may be linguistically negotiated
and aligned between conversational partners in dialogue
\cite{Dobnik:2014aa,Dobnik:2015aa,Dobnik:2016af}.
 
As mentioned earlier, spatial descriptions do not refer to the actual
objects in space but to conceptual geometric representations of these
objects, which may be points, lines, areas and volumes. The
representation depends on how we view the scene, for example
\emph{under the water} (water $\approx$ surface) and \emph{in the
  water} (water $\approx$ volume).  The influence of \emph{world
  knowledge} goes beyond object conceptualisation. Some prepositions
are more sensitive to the way the objects interact with each (their
dynamic kinematic routines) while other are more sensitive to
the way the objects relate geometrically \cite{CoventryEtAl:2001}.

Finally, because situated agents are located within dynamic linguistic
and perceptual environments they must continuously adapt their
understanding and representations relative to these context. On the language side they must maintain
language coordination with dialogue partners
\cite{Clark:1996aa,Fernandez:2011aa,schutte2017,Dobnik:2017aa}. A good
example of adaptation of contextual meaning through linguistic
interaction is the coordinated assignment of frame of reference
mentioned earlier.

In summary, the meaning of spatial descriptions is dynamic, dependent
on several sources of contextually provided knowledge which provide a
challenge for its computational modelling because of its contextual
underspecification and because it is difficult to provide and
integrate that kind of knowledge. On the other hand, a computational
system taking into account these meaning components in context would
be able to understand and generate better, more human-like, spatial
descriptions and engage in more efficient communication in the domain
of situated agents and humans. Furthermore, it could exploit the synergies
between different knowledge sources to compensate missing knowledge in
one source from another
\cite{SteelsLoetzsch:2009,Skocaj:2011fu,schutte2017}.

\subsection{Modular Mechanistic (Neural) Models of Spatial Language}

The discussion in the preceding section highlighted the numerous factors that impinge on the semantics of spatial language. It is this multiplicity of factors that make spatial language such a useful case study for this paper, the complexity of the problem invites a modular approach where the solution can be built in a piecewise manner and then integrated. One challenge to this approach to spatial language is the lack of an overarching theory explaining how these different factors should be integrated, examples of candidate theories that could act as a starting point here include \cite{Herskovits:1987:LSC:535779} and \cite{Coventry:2005ab}.

At the same time there are a number of examples of neural models in the literature that could provide a basis for the design of specific modules. We have already discussed \cite{regier:1996} which captured geometric factors and paths of motion. Another example of a mechanistic neural model of
spatial descriptions is described in \cite{Coventry:2005aa}. Their
system processes dynamic visual scenes containing three objects: a
teapot pouring water into a cup and the network learns to optimise,
for each temporal snapshot of a scene, the appropriateness score of a
spatial description obtained in subject experiments. The idea behind
these experiments is that descriptions such as \emph{over} and
\emph{above} are sensitive to a different degree to geometric and
functional properties of a scene, the latter arising from the
interactions between objects as mentioned earlier. The model is split
into three modules: (i) a vision processing module that deals with
detection of objects from image sequences that show the interaction of
objects, the tea pot, the water and the cup, using an attention
mechanism, (ii) an Elman recurrent network that learns the dynamics
of the attended objects in the scene over time, and (iii) a dual
feed-forward vision and language network to which representations from
the hidden layer of the Elman network are fed and which learns how to
predict the appropriateness score of each description for each
temporal configuration of objects. Each module of this network is
dedicated to a particular task: (i) to recognition of objects, (ii) to
follow motion of attended objects in time and (iii) to integration of
the attended object locations with language to predict the
appropriateness score, factors that have been identified to be
relevant for computational modelling of spatial language and cognition through previous experimental
work \cite{CoventryEtAl:2001}. The example shows
the effectiveness of representing networks as modules and their
possibility of joint training where individual modules
constrain each other.

The model could be extended in several ways. For example, contemporary
CNNs and RNNs could be used which have become standard in neural
modelling of vision and language due to their state-of-the-art
performance. Secondly, the approach is trained on a small dataset of
artificially generated images of a single interactive configuration of
three objects.\footnote{To be fair to the authors, their intention was not to build an image captioning system but to show that modular
  networks can optimise human experimental judgements.} An open
question is how the model scales on a large corpus of image
descriptions \cite{Krishna:2016aa} where considerable noise is added. There will be several objects, their appearance
and location may be distorted by the angle at which the image is
taken, there are no complete temporal sequences of objects and the
corpora typically does not contain human judgement scores on how
appropriate a description is given an image. Finally, Coventry et. al.'s 
model integrates three modalities used in spatial cognition, but as we
have seen there are several others \cite{Coventry:2005aa}. An important aspect is grounded
linguistic interaction and adaptation between agents. For example,
\cite{Lazaridou:2016aa} describe a system where two networks are
trained to perform referential games (dialogue games performed over
some visual scene) between two agents. In this context, the agents
develop their own language interactively. An open research question is
whether parameters such frame of reference intended by the speaker of a
description could also be learned this way. Note that this is not
always overtly specified, e.g. \emph{from my left}.

Sometimes a mechanistic design of the network architecture constrains what a model can learn in undesirable ways. For example, Kelleher \& Dobnik argue that contemporary image captioning networks as in Figure~\ref{fig:image-captioning-schematic} have been configured in a way that they capture visual properties of objects rather than spatial relations between them \cite{kelleherdobnik2017}. Consequently, within the captions generated by these systems the relation between the preposition and the object is not grounded in geometric representation of space but only in the linguistic sequences through the decoder language model where the co-occurrence of particular words in a sequence is estimated. \cite{Dobnik:2013aa,Dobnik:2014ab} show that a language model is predictive of functional relations between objects that spatial relations are also sensitive to but in this case the geometric dimension is missing. This indicates that the architecture of these image-captioning systems, although modular, ignores important domain theoretic considerations and hence are best understood as close to the phenomenological (black-box) than the mechanistic (grey-box) network design philosophy this paper advocates.

In summary, it follows that an appropriate computational model of spatial language
should consist of several connected modalities (for which individual
neural network architectures are specified) but also of a general
network that connects these modalities, thus akin to the specialised
regions and their interconnections in the brain
\cite{Roelofs:2014aa}. The challenge of creating and training such a system is obviously significant, however one feature of neural network training that may make this task easier is that it is possible to back-propagate through a pre-trained network. This opens the possibility of pre-training networks as modules (sometimes even on different datasets) that carry out specific theory-informed tasks and then training larger systems that represent the full-theory by including these pre-trained modules components within the system and training other modules and/or integration layers while keeping the weights of the pre-trained modules frozen during training.




\section{Conclusion and Future Research}\label{sec:future}

DNNs provide a platform for machine learning that permits great
flexibility in combining top-down specification (in terms of
hand-designed structures and rules) and data driven
approaches. Designers can tailor the network structures to each
individual learning problem and therefore effectively reach the goal
of combining mechanistic and phenomenological approaches: a problem that has
been investigated in NLP for several decades. The strength of DNNs is in
the compositionality of perceptrons or neural units, and indeed
networks themselves, which represent individual classification
functions that can be combined in novel ways. This was not possible
with other approaches in machine learning to the same degree with a consequences that these worked more as black boxes. Finally,
although we are not advocating that there is a direct similarity between DNNs
and human cognition, it is nonetheless the case that DNNs are inspired by
neurons and connectionist organisation of human brain and hence at
some high abstract level they share some similarities, for example basic classification units combine to larger structures, the 
structures get specialised to modules to perform certain tasks, and 
training and classification is performed across several
modules. Therefore, this might be a possible explanation that DNNs have been so
successful in computational modelling of language and vision, the surface manifestations of the underlying human cognition, as at some abstract level they represent a similar architecture to human cognition.


\section*{Acknowledgements}

The research of Dobnik was supported by a grant from the Swedish
Research Council (VR project 2014-39) for the establishment of the
Centre for Linguistic Theory and Studies in Probability (CLASP) at
Department of Philosophy, Linguistics and Theory of Science (FLoV),
University of Gothenburg.

The research of Kelleher was supported by the ADAPT Research
Centre. The ADAPT Centre for Digital Content Technology is funded
under the SFI Research Centres Programme (Grant 13/RC/2106) and is
co-funded under the European Regional Development Funds.

\bibliography{laml-2017-dobnik-kelleher} 
\bibliographystyle{plain}

\end{document}